%% file: main.tex
\documentclass[10pt,twocolumn,letterpaper]{article}

\usepackage[pagenumbers]{iccv} % To force page numbers, e.g. for an arXiv version


\definecolor{iccvblue}{rgb}{0.21,0.49,0.74}
\usepackage[pagebackref,breaklinks,colorlinks,allcolors=iccvblue]{hyperref}
\usepackage{cuted}
\usepackage{multirow}
\usepackage{xcolor,colortbl}
\definecolor{gainsboro}{rgb}{0.86, 0.86, 0.86}

\usepackage[accsupp]{axessibility}  % Improves PDF readability for those with disabilities.

\def\paperID{14271} % *** Enter the Paper ID here
\def\confName{ICCV}
\def\confYear{2025}

\title{Know Your Attention Maps: Class-specific Token Masking for \\ Weakly Supervised Semantic Segmentation}

\author{
Joëlle Hanna \quad Damian Borth \\
University of St.Gallen, Switzerland \\
{\tt\small \{joelle.hanna, damian.borth\}@unisg.ch}
}

\begin{document}
\maketitle
\input{0_abstract}
\input{1_intro}
\input{2_relatedworks}
\input{3_approach}
\input{4_datasets}
\input{5_experiments}
\input{6_discussion}
\input{7_conclusion}
{
    \small
    \bibliographystyle{ieeenat_fullname}
    \bibliography{main}
}

\end{document}

%% file: 0_abstract.tex
\begin{abstract}
Weakly Supervised Semantic Segmentation (WSSS) is a challenging problem that has been extensively studied in recent years. Traditional approaches often rely on external modules like Class Activation Maps \cite{Zhou2016LearningDF} to highlight regions of interest and generate pseudo segmentation masks. In this work, we propose an end-to-end method that directly utilizes the attention maps learned by a Vision Transformer (ViT) \cite{Dosovitskiy2020AnII} for WSSS. We propose training a sparse ViT with multiple \texttt{[CLS]} tokens (one for each class), using a random masking strategy to promote \texttt{[CLS]} token - class assignment. At inference time, we aggregate the different self-attention maps of each \texttt{[CLS]} token corresponding to the predicted labels to generate pseudo segmentation masks\footnote{Code is available at \href{https://github.com/HSG-AIML/TokenMasking-WSSS}{github.com/HSG-AIML/TokenMasking-WSSS}}. Our proposed approach enhances the interpretability of self-attention maps and ensures accurate class assignments. Extensive experiments on two standard benchmarks and three specialized datasets demonstrate that our method generates accurate pseudo-masks, outperforming related works. Those pseudo-masks can be used to train a segmentation model which achieves results comparable to fully-supervised models, significantly reducing the need for fine-grained labeled data.
\end{abstract}

%% file: 1_intro.tex
\section{Introduction}
\label{sec:intro}
Semantic segmentation is a fundamental task in computer vision, aiming to assign a class label to each pixel in an image.  This fine-grained understanding of visual data is essential for various domain-specific applications, such as medical imaging, autonomous driving, and remote sensing. Traditionally, achieving high performance in semantic segmentation has relied heavily on the availability of large annotated datasets. However, acquiring such fine-grained labels is often labor-intensive, expensive or even impractical, especially in specialized domains where expert knowledge is required.
% \\\\
\begin{figure}[t]
    \centering
 \includegraphics[scale=0.24]{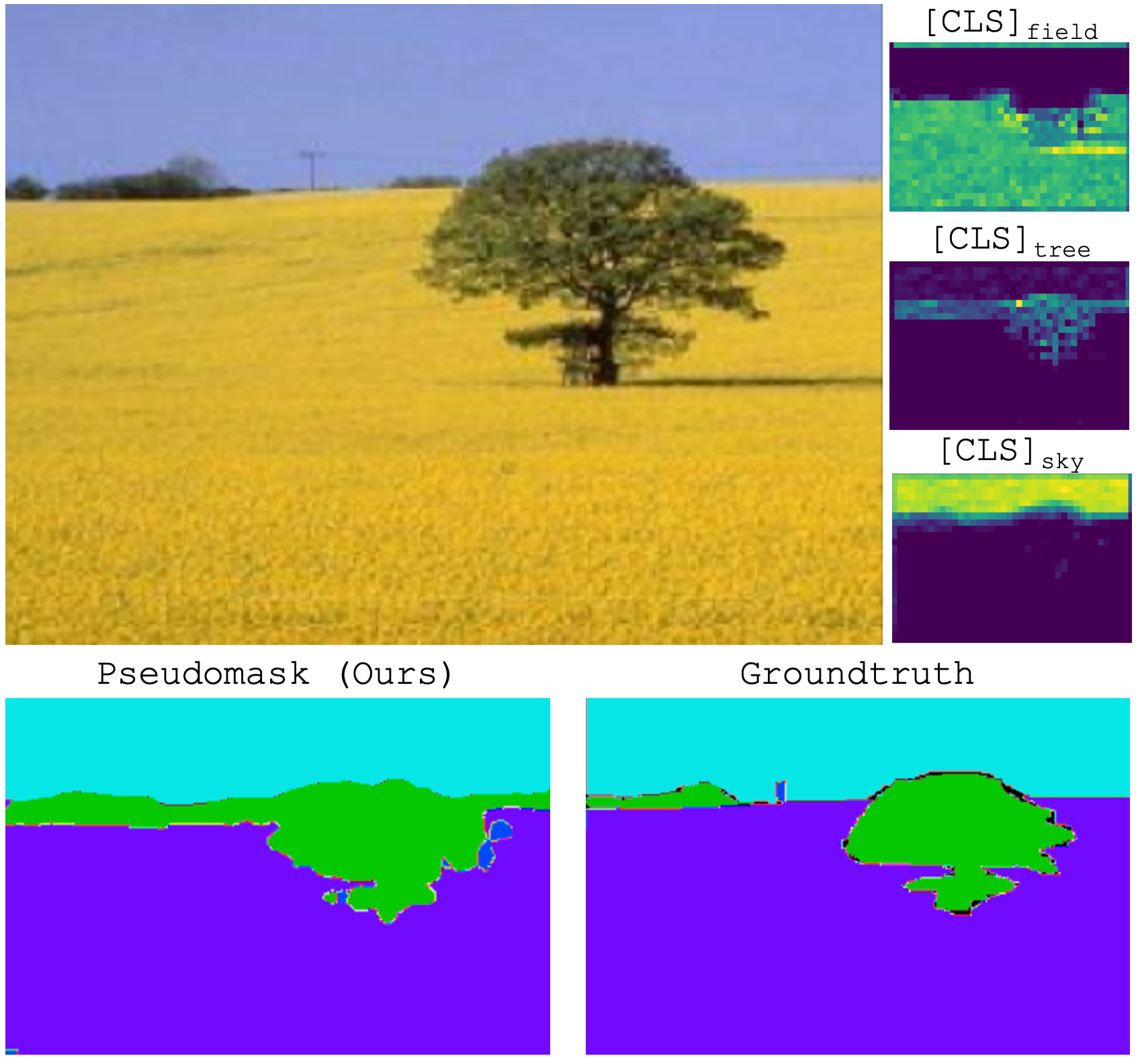}
    \caption{Our method uses multiple \texttt{[CLS]} tokens in ViTs to encourage class-specific self-attention maps. By aggregating these maps, we generate a pseudo-mask (bottom left) comparable to the ground truth (bottom right), without ever using the fine-grained segmentation labels.}
    \label{fig:front-pager}
    \vspace{-2.0em}
\end{figure}
\\
\indent
To address these challenges, weakly supervised semantic segmentation (WSSS) has emerged as a promising alternative. WSSS leverages weaker forms of supervision, such as image-level labels, to reduce the burden of detailed annotations. A common approach in WSSS involves the use of Class Activation Maps \cite{Zhou2015LearningDF}, which highlight the most discriminative regions of an image for a particular class. Despite their effectiveness, CAMs suffer from limitations such as coarse localization and an inability to capture fine details, as they only emphasize the most prominent features. Moreover, generating CAMs requires an additional module on top of the existing model, adding complexity to process.
% \\\\
% \noindent
\\
Given these limitations, an alternative approach is to leverage the self-attention maps of Vision Transformers (ViTs,\cite{Dosovitskiy2020AnII}).  Unlike CAMs, which rely on additional modules, the attention mechanism in ViTs inherently provides interpretability by capturing spatial dependencies across the entire image. Vision Transformers, with their self-attention mechanisms, have shown remarkable performance in various vision tasks. However, a significant challenge in using ViTs for WSSS is the inability to assign specific classes to individual attention heads, making it difficult to interpret and utilize their outputs for segmentation tasks effectively \cite{Hanna2023SparseMV}.
% \\\\
%  \noindent
% \\ \indent
\\ \indent
In this work, we propose a novel approach for WSSS that leverages the multi-head self-attention mechanism of ViTs in a more structured manner. Instead of relying on a single \texttt{[CLS]} token that aggregates global information, we introduce multiple \texttt{[CLS]} tokens, each designed to correspond to a specific class in the segmentation task. This allows the model to decompose the scene into distinct class-specific representations. Ideally, each \texttt{[CLS]} token would correspond to one class. However, there is no inherent guarantee that each \texttt{[CLS]} token will only capture information relevant to its intended class. To address this, we implement a novel masking mechanism on the \texttt{[CLS]} token output embeddings to help ensure that each \texttt{[CLS]} token is properly assigned to a single class (Figure \ref{fig:front-pager}).
% \\\\
\\ \indent
Furthermore, since not all attention heads contribute equally to class-specific feature extraction, we introduce a pruning mechanism to remove redundant or noisy heads. By selectively removing less important heads, we aim to sharpen the attention mechanism, making the remaining heads more interpretable. The ultimate goal is to leverage these refined attention heads to create accurate class-specific segmentation masks, thus enabling the ViT to perform weakly supervised semantic segmentation effectively.
\\\\
This work aims to bridge the gap between the powerful capabilities of ViTs and the need for class-specific feature embeddings in complex vision tasks. The contributions of this work can be summarized as follows:

\begin{itemize}\itemsep0em
 \item We propose to use multiple \texttt{[CLS]} tokens instead of one, and develop a masking mechanism for the \texttt{[CLS]} tokens that reliably assigns each \texttt{[CLS]} token to its respective class.
\item We introduce attention head pruning during training, significantly reducing noise in self-attention maps and improving pseudo-mask quality.
\item We validate the application of this method in WSSS on two standard datasets and across three distinct domains: remote sensing, medical imaging, and general scene understanding. Each domain presents unique challenges: remote sensing requires accurate parsing of complex landscapes, medical imaging demands precise object-centric segmentation for diagnostic purposes, and general scene understanding involves diverse and sparse labels.
% \item In comparison with fully-supervised segmentation setups, we show that our weakly-supervised approach achieves similar performance while relying only on image-level labels.
\item We demonstrate that our work can effectively reduce the need for fine-grained labeled data, benefiting both small-scale and large-scale datasets.
\end{itemize}

%% file: 2_relatedworks.tex
\section{Related Works}
\label{sec:related_works}

\subsection{Weakly Supervised Semantic Segmentation}

Weakly supervised semantic segmentation (WSSS) offers an alternative to traditional supervised methods by relying on less precise forms of supervision, such as image-level labels \cite{Chen2022ClassRM, Wang2020SelfSupervisedEA}, bounding boxes \cite{Khoreva2017SimpleDI, Song2019BoxDrivenCR}, or scribbles \cite{Lin2016ScribbleSupSC, Unal2022ScribbleSupervisedLS}. WSSS approaches can be broadly categorized into single-stage and multi-stage methods, each with distinct workflows and objectives.

Single-stage WSSS methods aim to directly generate segmentation maps from weak labels without intermediate processing. These approaches often leverage end-to-end models that integrate weak supervision signals with advanced architectures. Recent advancements include Vision Transformers (ViTs) specifically tailored for WSSS. Hanna \etal \cite{Hanna2023SparseMV} introduced a multimodal ViT enforcing sparsity in attention heads during training, enabling the direct generation of high-quality segmentation maps. Similarly, Xu \etal \cite{Xu2022MulticlassTT} proposed a Multi-Class Token Transformer that employs class tokens to enhance localization for multiple objects in a single forward pass. These single-stage approaches simplify the WSSS pipeline by bypassing the need for pseudo-mask refinement stages, making them computationally efficient and suitable for real-time applications.

Multi-stage WSSS methods, on the other hand, involve an iterative process where intermediate representations, such as pseudo segmentation masks (pseudo-masks), are generated and refined before final segmentation. This process typically begins with generating coarse localization maps from weak labels, followed by refinement and re-training in a fully supervised manner. One common approach is based on Class Activation Maps (CAMs), initially proposed by Zhou \etal \cite{Zhou2016LearningDF}, which provide rough localization cues by highlighting discriminative image regions. Building on CAMs, Ahn \etal \cite{Ahn2018LearningPS} introduced AffinityNet, which refines these maps by considering pixel affinities to propagate segmentation information. Wei \etal \cite{Wei2017ObjectRM} extended this idea with Adversarial Erasing, iteratively suppressing confident CAM regions to uncover complete object extents.

Further advancements in multi-stage WSSS have integrated various forms of weak supervision and iterative refinement strategies. For instance, Chen \etal \cite{Chen2021SemiSupervisedSS} proposed a dual-network approach, where a segmentation network and a pseudo-label generation network are alternately trained to progressively improve pseudo-mask quality. Additionally, recent works \cite{Kweon_2024_CVPR, Chen2023SegmentAM} have leveraged the Segment Anything Model (SAM) \cite{Kirillov2023SegmentA} during inference, employing post-processing techniques to refine segmentation results.

\subsection{Attention Maps of Vision Transformers}
Many studies have recently demonstrated that attention maps generated by Vision Transformers are not only effective for downstream tasks but also offer insights into the decision-making processes of these models. Caron \etal \cite{Caron2021EmergingPI} introduced a self-supervised learning method called DINO, which significantly enhances the interpretability of ViTs. They observed that self-supervised ViTs can automatically learn class-specific features and produce unsupervised object segmentations. Furthermore, Darcet \etal \cite{Darcet2023VisionTN} identified artifacts in the attention maps of both supervised and self-supervised ViT networks. By introducing register tokens, they provided an effective solution to eliminate these artifacts, leading to more interpretable attention maps.

%% file: 3_approach.tex
\section{Approach}
\label{sec:approach}

\begin{figure*}
\centering
\begin{subfigure}{.5\textwidth}
  \centering
  \includegraphics[width=\linewidth]{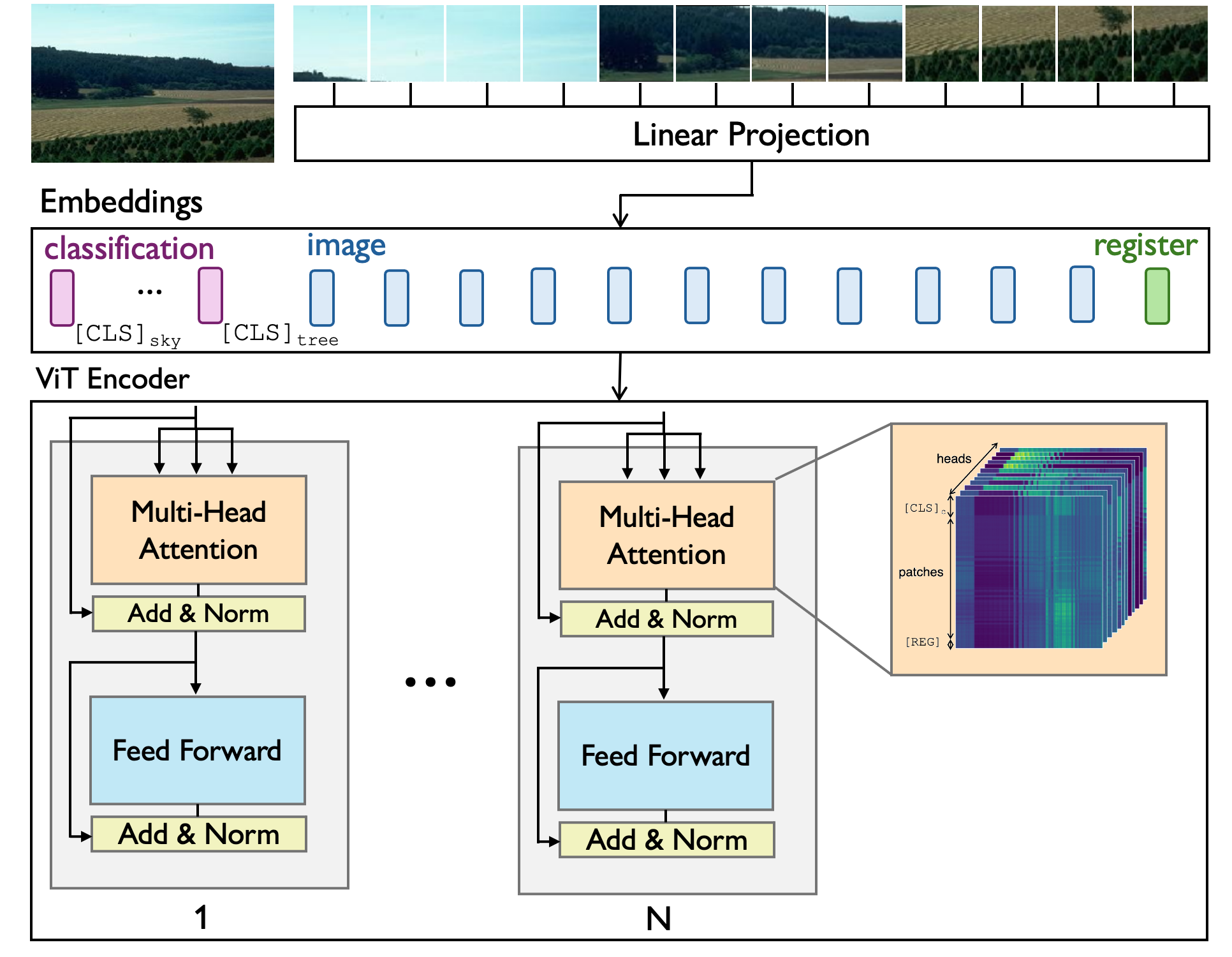}
  \caption{Overview of the method's architecture.}
  \label{fig:sub1}
\end{subfigure}%
\begin{subfigure}{.5\textwidth}
  \centering
  \includegraphics[width=0.85\linewidth]{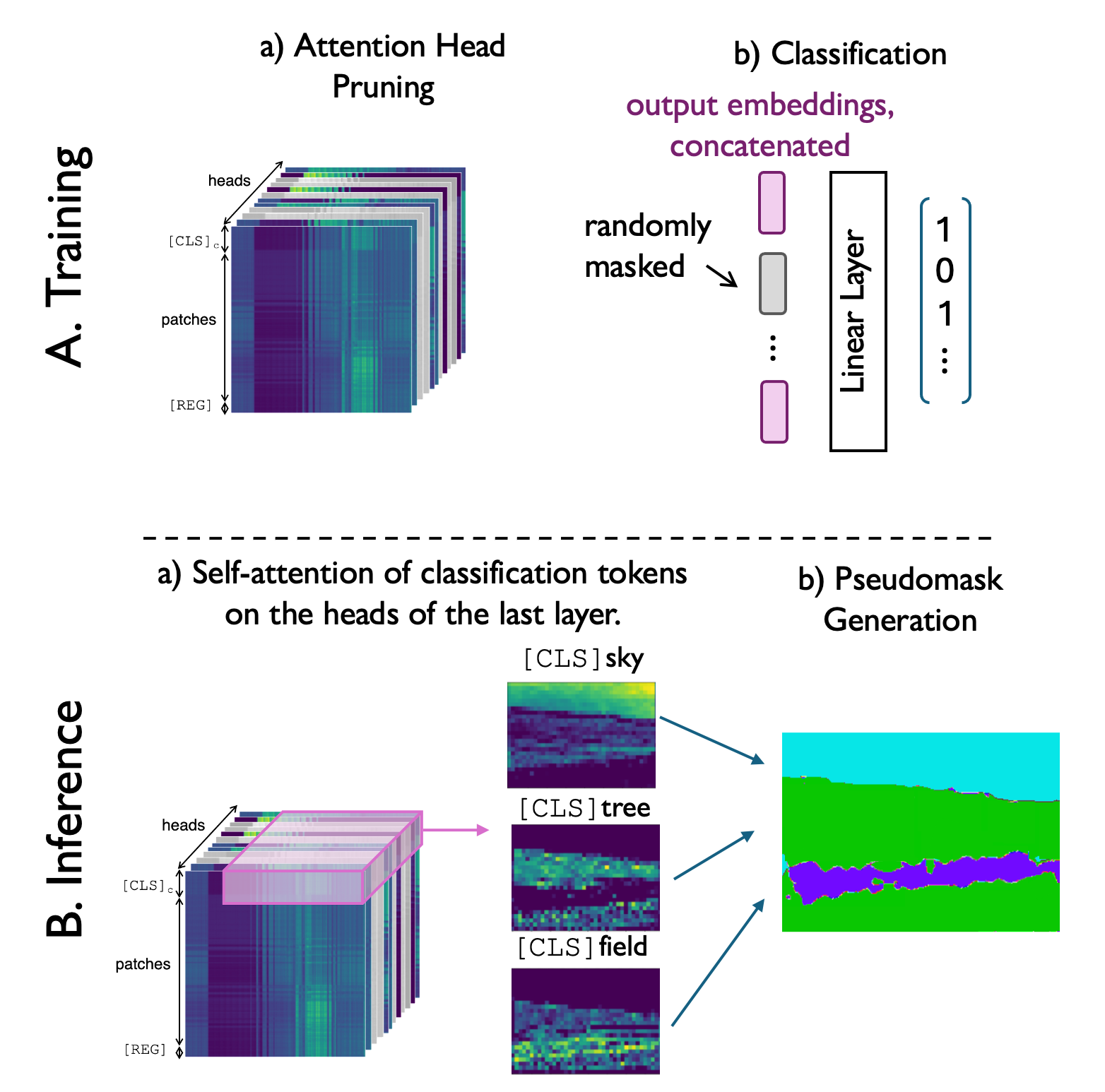}
  \caption{Training and inference pipelines}
  \label{fig:sub2}
\end{subfigure}
\caption{Our proposed approach divides the input image into patches, which are projected into a sequence of image tokens. To this sequence, we append C classification tokens (\texttt{[CLS]}) and a register token \texttt{[REG]}. During training, attention heads are pruned to enhance the interpretability of the self-attention maps, and the output embeddings of the \texttt{[CLS]} tokens are randomly masked before classification to enforce class assignments. During inference, the self-attention of the \texttt{[CLS]} tokens corresponding to the predicted labels are reshaped into the input image dimensions and used to generate pseudo-masks.
}
\vspace{-1em}
\label{fig:method_diagram}
\end{figure*}

In this section, we describe our approach and illustrate it in Figure \ref{fig:method_diagram}.

\subsection{Vision Transformer Basics}

Given an image $\mathbf{x} \in \mathbb{R}^{H \times W \times C}$, it is divided into $N$ patches, each of size $P \times P$. These patches are flattened and embedded into a higher-dimensional space:

\[ \mathbf{z}_0 = [\mathbf{x}_p^1 \mathbf{E}; \mathbf{x}_p^2 \mathbf{E}; \ldots; \mathbf{x}_p^N \mathbf{E}] + \mathbf{E}_{\text{pos}} \]

where $\mathbf{E}$ is the embedding matrix and $\mathbf{E}_{\text{pos}}$ represents positional encodings. A special \texttt{[CLS]} token is prepended to the sequence of embedded patches:
\vspace{-0.5em}
\[ \mathbf{z}_0 = [\texttt{[CLS]}; \mathbf{z}_0] \]

The sequence is then processed through Transformer encoder layers, leveraging multi-head self-attention (MSA) and feed-forward neural networks to generate feature representations \cite{Vaswani2017AttentionIA}:

\[ \mathbf{z}_\ell' = \text{MSA}(\text{LN}(\mathbf{z}_{\ell-1})) + \mathbf{z}_{\ell-1} \]
\[ \mathbf{z}_\ell = \text{MLP}(\text{LN}(\mathbf{z}_\ell')) + \mathbf{z}_\ell' \]

where $\mathbf{z}_\ell$ is the output of the $\ell$-th layer, and $\text{LN}$ denotes layer normalization. The final output representation corresponding to the \texttt{[CLS]} token is used for classification tasks:
\vspace{-0.5em}
\[ \mathbf{y} = \text{MLP}(\mathbf{z}_L^{\texttt{[CLS]}}) \]

where $\mathbf{z}_L^{\texttt{[CLS]}}$ is the output of the final Transformer layer corresponding to the \texttt{[CLS]} token.

\subsection{ViT with Multiple \textbf{\texttt{[CLS]}} Tokens}
\label{sec:method:multiple_tokens}

We extend the standard ViT architecture, which typically uses a single \texttt{[CLS]} token, to incorporate multiple \texttt{[CLS]} tokens. Each \texttt{[CLS]} token is intended to represent a different class in a multi-label classification setup. This enables the model to handle images with multiple objects and infer segmentation masks for each class separately. The extended embedding sequence is given by:
\vspace{-0.5em}
\begin{equation*}
    \mathbf{z}_0 =[\texttt{[CLS]}_1; \texttt{[CLS]}_2; \ldots; \texttt{[CLS]}_C; \mathbf{z}_0]
\end{equation*}

where $C$ is the number of classes. Each \texttt{[CLS]} token's output $\mathbf{z}_L^{\texttt{[CLS]}_c}$ for class $c$ is processed to generate class-specific predictions.

\paragraph{Additional Register Token}
Following the work from Darcet \etal \cite{Darcet2023VisionTN}, we append an additional learnable token that the model can use as a register, \texttt{[REG]}, to the end of the token sequence. The register token captures general context to prevent it from 'polluting' the \texttt{[CLS]} self-attention.
\vspace{-0.5em}
\[ \mathbf{z}_0 = [\texttt{[CLS]}_1; \texttt{[CLS]}_2; \ldots; \texttt{[CLS]}_C; \mathbf{z}_0; \texttt{[REG]}] \]

\subsection{Random Masking of \textbf{\texttt{[CLS]}} Tokens}
\label{sec:method:random_masking}

To address the issue of the lack of hard assignment between each \texttt{[CLS]} token and its corresponding class, we implement a \texttt{[CLS]} masking strategy during training. This approach aims to reduce interference between classes, as the model would learn to rely on the available tokens to make class-specific decisions, promoting more robust class representations.
The process is as follows:

\begin{enumerate}
    \item \textbf{Random Selection}: During each training iteration, a subset of \texttt{[CLS]} tokens that do not correspond to the current image labels are randomly selected to be masked. The masking ratio is set to 50\%, with a sensitivity analysis of this parameter presented in Section \ref{sec:exp:sensitivity_analysis}. Let \( \mathcal{Y} \) be the set of true labels for the image $\mathbf{x}$. We define the masking function \( m(i) \) for \texttt{[CLS]} token \( \text{CLS}_i \) as follows:
    \vspace{-0.2cm}
    \[
    m(i) =
    \begin{cases}
    0 & \text{if } i \in \mathcal{Y} \\
    1 & \text{if } i \notin \mathcal{Y} \text{ and randomly selected to be masked}
    \end{cases}
    \]

    \item \textbf{Output Embedding Masking}: To ensure masked \texttt{[CLS]} tokens do not contribute to the final predictions, we mask their output embeddings:

    \[
    \mathbf{z}_L^{\texttt{[CLS]}_i} = \mathbf{z}_L^{\texttt{[CLS]}_i} \cdot (1 - m(i))
    \]

    where \( \mathbf{z}_L^{\texttt{[CLS]}_i} \) is the final output embedding for the \(\texttt{[CLS]}_i\) token.
\end{enumerate}

\subsection{Sparse Training of the Vision Transformer}
\label{sec:method:sparse_training}
Aside from the challenge of assigning each \texttt{[CLS]} token to a specific class, another essential task is making the self-attention of these \texttt{[CLS]} tokens interpretable, providing clear and detailed shapes of different classes. Following common findings that many heads can be pruned in a transformer \cite{Voita2019AnalyzingMS, Hanna2023SparseMV}, our solution is to prune redundant heads that might introduce noise, leading to clearer and more accurate class-specific representations.

To achieve this, we modify the ViT architecture by incorporating gating units \( g_i \) into each attention head. Those scalar gates are multiplied by the output of each head \( i \). This  modifies the multi-head self-attention (MSA) mechanism as follows:

\[
\text{MSA}(Q, K, V) = \text{concat}_i(g_i \cdot \text{head}_i) W^O
\]

Here, $W^O$ is a  parameter matrix. Our objective is to entirely disable the less significant heads. While \( L_0 \) regularization would be ideal for this purpose \cite{Louizos2017LearningSN}, it is non-differentiable. To address this, we employ a stochastic relaxation method that uses Hard Concrete distributions \cite{Maddison2016TheCD}. By summing the probabilities of heads being non-zero, we can approximate the \( L_0 \)  norm:

\[
\mathcal{L}_{\text{reg}} = \sum_i (1 - P(g_i = 0 \mid \phi_i))
\]

Here, \( \phi_i \) are the parameters of the Hard Concrete distribution. The training objective becomes:

\begin{equation}
    \mathcal{L} = \mathcal{L}_{\text{cls}} + \lambda \mathcal{L}_{\text{reg}}
    \label{eq:loss}
\end{equation}

\( \mathcal{L}_{\text{cls}} \) is the  loss used for training the classifier, e.g. binary cross-entropy. The parameter $\lambda$ is used to control the number of heads to be pruned. It is set to $0.01$, resulting in approximately two-thirds of the heads being pruned. A sensitivity analysis on that parameter is presented in the Supplementary Material.

\subsection{Segmentation Pseudo-mask Generation}
\label{sec:method:pseudo-mask_generation}

The final step in our approach is to generate pseudo segmentation masks by combining the self-attention maps for each predicted class from the corresponding \texttt{[CLS]} tokens. For each image \( \mathbf{x} \in \mathbb{R}^{H \times W \times C}\), we extract the representations from the \texttt{[CLS]} tokens corresponding to the predicted classes, each denoted as \(Z \in \mathbb{R}^{1 \times \frac{H \times W}{P^2}}\). These representations are then reshaped to align with the spatial dimensions of the image, resulting in attention scores of dimensions \(\mathbb{R}^{\frac{H}{P} \times \frac{W}{P}}\). Then, we threshold each self-attention map of the \texttt{[CLS]} tokens for the predicted classes to binarize them. We then combine all binarized maps, starting with the class of the lowest probability (logit) and ending with the highest. Any pixel not assigned to a class is replaced by the most common value of neighboring pixels.

%% file: 4_datasets.tex
\section{Datasets}
\label{sec:datasets}

We evaluate our approach on two standard datasets: MS COCO 2014 \cite{Lin2014MicrosoftCC} and Pascal VOC 2012 \cite{Everingham2010ThePV} augmented by the SBD dataset \cite{Hariharan2011SemanticCF}, following common practices. We also use three domain-specific datasets: the DFC2020 remote sensing dataset \cite{rha7-m332-19}, the EndoTect Polyp Segmentation Medical Imaging dataset \cite{Hicks2020TheE2}, and the ADE20K general scene parsing dataset \cite{Zhou2017ScenePT}.

\vspace{-0.5em}
\paragraph{The DFC2020 dataset} includes $\sim$5K paired Sentinel 1 (SAR) and Sentinel 2 (multispectral) satellite observations for training, and $\sim$1K for testing, each image sized 256 × 256. It provides pixel-level land cover annotations for eight imbalanced classes. Image-level labels are generated for training, keeping classes covering $\geq$ 10\% of the image.
\vspace{-0.5em}
\paragraph{The EndoTect Polyp Segmentation dataset} consists of around 110K high-resolution endoscopic images with detailed polyp region annotations. Polyp segmentation is crucial for the early diagnosis of cancer, making this dataset important for evaluating medical imaging tasks with high variability in polyp appearance.
\vspace{-0.5em}
\paragraph{The ADE20K dataset} contains images annotated with 150 object categories from various scenes, including urban, indoor, and natural environments, making it suitable for evaluating generalization to real-world scenarios.
\\\\
Across all datasets, we use classification labels during training and pixel-level annotations to evaluate the weak segmentation accuracy of our method.

%% file: 5_experiments.tex
\section{Experiments}
\label{sec:experiments}

\subsection{Models Architecture}
We use a ViT-B \cite{Dosovitskiy2020AnII} as our base model with a fully connected layer for classification, to generate pseudo-masks as initial segmentation predictions. These pseudo-masks are then used to train a UNet model \cite{Ronneberger2015UNetCN}, which serves as our segmentation model, to produce the final results.

\subsection{Pseudo-masks Evaluation}
\begin{figure}
    \centering
    \includegraphics[width=\linewidth]{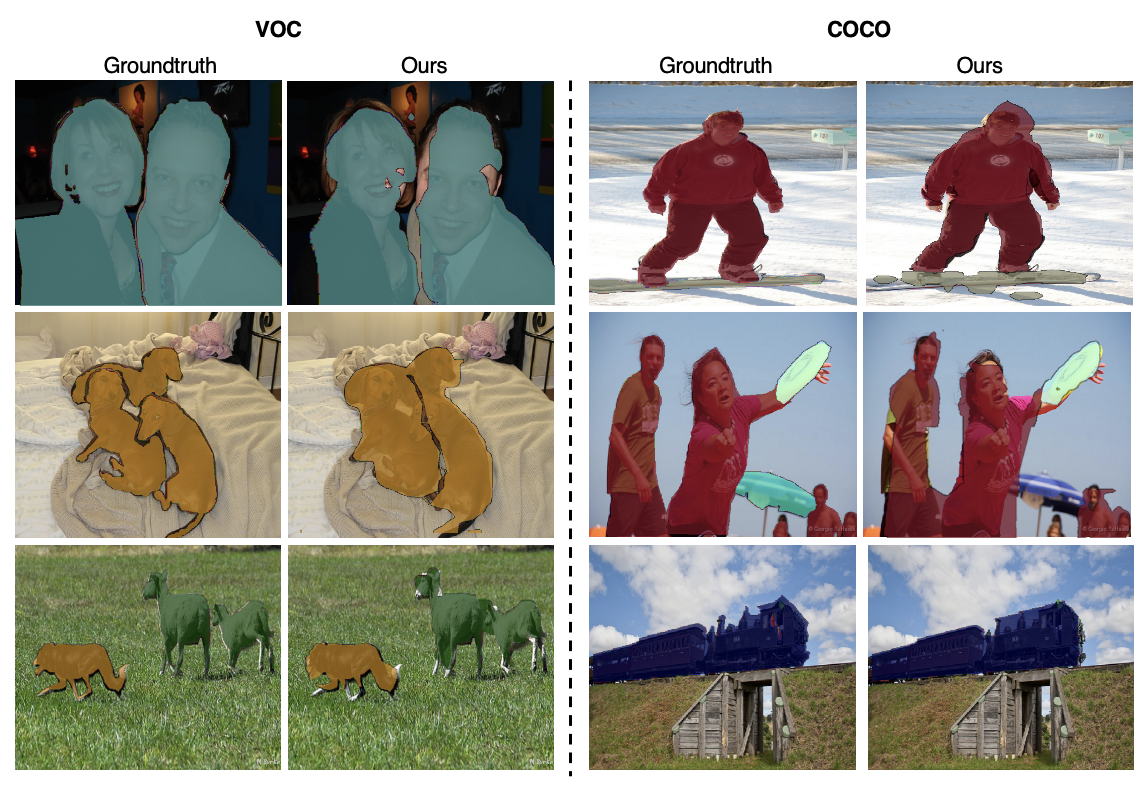}
    \caption{Pseudo-mask results on the Pascal VOC and MS COCO datasets.}
    \label{fig:voc_coco_comp}
    \vspace{-1em}
\end{figure}

We evaluate the quality of the pseudo-masks generated by our method by comparing their accuracy against those from existing state-of-the-art WSSS approaches and report the results in terms of mean Intersection over Union (mIoU). Table \ref{tab:baselines} shows the quantitative performance of our approach on the Pascal VOC 2012 \texttt{train} and \texttt{val} sets. We note that single-stage methods, which generate pseudomasks in a single step, achieve competitive results with simpler pipelines than multi-stage approaches. Among these, our method achieves the best validation mIoU of 73.7\%, slightly outperforming the previous best single-stage method, DuPL \cite{Wu2024DuPLDS} (73.5\%). Moreover, visually, as shown in Figure \ref{fig:voc_coco_comp}, the pseudo-masks generated by our method closely match the ground truth segmentation masks on the Pascal VOC and MS COCO datasets.

\begin{table}
    \centering
    \caption{Evaluation of pseudo-masks. The results are evaluated on the VOC train and val sets, and reported in mIoU (\%). \dag denotes using ImageNet-21k pretraining. }
    \scalebox{0.93}{
    \begin{tabular}{cccc}
        \toprule
       \textbf{Method} & \textbf{Backbone} & \texttt{train} & \texttt{val} \\
        \toprule
        \multicolumn{4}{l}{\textbf{Multi-stage WSSS methods}} \\
        % ViT-PCM + CRF \cite{Rossetti2022MaxPW} \tiny ECCV 2022 & ViT-B^{\tiny{\dag}}  & 71.4  & 69.3 \\
        ViT-PCM + CRF \cite{Rossetti2022MaxPW} \tiny ECCV 2022 & ViT-B\textsuperscript{\dag}  & 71.4  & 69.3 \\
        ReCAM \cite{Chen2022ClassRM} \tiny CVPR 2022 & ResNet50 & 70.5 & -  \\
        I/C-CTI \cite{Yoon2024ClassTI} \tiny CVPR 2024 & DeiT-S & 73.7 & - \\
        \midrule
        \multicolumn{4}{l}{\textbf{Single-stage WSSS methods}} \\
        % ViT-PCM \cite{Rossetti2022MaxPW} \tiny ECCV 2022 & ViT-B^{\tiny{\dag}}  & 67.7  & 66.0 \\
        ViT-PCM \cite{Rossetti2022MaxPW} \tiny ECCV 2022 & ViT-B\textsuperscript{\dag}  & 67.7  & 66.0 \\
        AFA \cite{Ru2022LearningAF} \tiny CVPR 2022 & MiT-B1 & 68.7 & 66.5   \\
        ToCo \cite{Ru2023TokenCF} \tiny CVPR 2023 & ViT-B & 72.2 & 70.5   \\
        DuPL \cite{Wu2024DuPLDS} \tiny CVPR 2024 & ViT-B & \textbf{75.1} & 73.5   \\
        \rowcolor{gainsboro}
        Ours & ViT-B & 74.5 & \textbf{73.7}  \\
         \bottomrule
    \end{tabular}}
    \label{tab:baselines}
    % \vspace{-1.2em}
\end{table}

\subsection{Final Segmentation Evaluation}
Table \ref{tab:baselines_seg} compares semantic segmentation results of different Weakly Supervised Semantic Segmentation (WSSS) methods on the VOC validation, VOC test, and COCO validation datasets, measured in mIoU. Multistage methods, such as I/C-CTI \cite{Yoon2024ClassTI}, achieve strong results, with the best VOC validation mIoU (74.1\%). Our method achieves the highest VOC test mIoU (73.5\%) and performs strongly on VOC validation (72.7\%) and COCO validation (43.2\%), highlighting its efficiency and generalization capabilities with a simplified pipeline. Additionally, as shown in Figure 4, the model performs well on specialized datasets like ADE20K and DFC2020, demonstrating its ability to handle diverse tasks. On ADE20K, it produces clear object boundaries in complex scenes, while on DFC2020, it segments detailed land-cover features effectively, even with imbalanced classes. These results further validate the robustness and adaptability of our method across various datasets and tasks.

\begin{table}[b]
    \centering
    \caption{Semantic Segmentation Results. The results are evaluated on the VOC val and test sets, and COCO val set, and reported in mIoU (\%). \dag denotes using ImageNet-21k pretraining.}
    \scalebox{0.85}{
    \begin{tabular}{ccccc}
        \toprule
       \multirow{2}{*}{\textbf{Method}} & \multirow{2}{*}{\textbf{Backbone}} & \multicolumn{2}{c}{\textbf{VOC}} & \textbf{COCO} \\
       & & \texttt{val} & \texttt{test} & \texttt{val} \\
        \toprule
        \multicolumn{5}{l}{\textbf{Multi-stage WSSS methods}} \\
        ReCAM \cite{Chen2022ClassRM} \tiny CVPR 2022 & DL-V2 & 68.4 & 68.2 & 45.0 \\
        % ViT-PCM \cite{Rossetti2022MaxPW} \tiny ECCV 2022 & ViT-B^{\tiny{\dag}}  & xxx & xxx & xxx \\
        MCTformer \cite{Xu2022MulticlassTT} \tiny CVPR 2022 & WR-38 & 71.9 & 71.6 & 42.0 \\
        I/C-CTI \cite{Yoon2024ClassTI} \tiny CVPR 2024 & ResNet38 & \textbf{74.1} & 73.2 & 45.4 \\
        \midrule
        \multicolumn{5}{l}{\textbf{Single-stage WSSS methods}} \\
        AFA \cite{Ru2022LearningAF} \tiny CVPR 2022 & MiT-B1 & 66.0 & 66.3 & 38.9 \\
        ToCo \cite{Ru2023TokenCF} \tiny CVPR 2023 & ViT-B & 69.8 & 70.5 & 41.3 \\
        DuPL \cite{Wu2024DuPLDS} \tiny CVPR 2024 & ViT-B & 72.2 & 71.6 & 43.6 \\
        \rowcolor{gainsboro}
        Ours & ViT-B & 72.7 & \textbf{73.5} & 43.2 \\
         \bottomrule
    \end{tabular}}
    \label{tab:baselines_seg}
    % \vspace{-1.2em}
\end{table}

\subsection{Specialized Datasets}
Table \ref{tab:baselines_specialized} shows the performance of our method on specialized datasets (DFC2020, Endotect, and ADE20K). Unlike the standard benchmarks, specialized datasets often contain unique characteristics such as imbalanced classes, fine-grained details, or domain-specific variations, making segmentation more difficult. Our method achieves state-of-the-art results among weakly supervised approaches on all specialized datasets, even surpassing fully supervised methods on DFC2020. This demonstrates its robustness and adaptability to real-world scenarios where labeled data is scarce.  A qualitative comparison for the three datasets is available in the Supplementary Material.

\begin{table}[t]
    \centering
    \caption{Performance on specialized datasets, showing fully supervised semantic segmentation methods and weakly supervised semantic segmentation methods. For each dataset we report \textit{Pixel Accuracy} and \textit{mIoU}. The top model overall is underlined, and the best WSSS model is bolded.}
    \scalebox{1}{
    \begin{tabular}{cccc}
       & & \textbf{Pixel Accuracy} & \textbf{mIoU} \\
        \toprule
        & \multicolumn{3}{l}{\textbf{Supervised Methods}} \\
         \multirow{7}{*}{\rotatebox[origin=c]{90}{\small DFC2020}} &  \small UNet  & 57.2 & 53.1 \\
         & \small ViT + seg.  & 46.4 & 43.4 \\
         & \small Robinson \etal \cite{9369830} \tiny 2021 & 70 & - \\
         \cmidrule(lr){2-4}
         & \multicolumn{3}{l}{\textbf{Weakly Supervised Methods}} \\
         &  \small ViT + GradCAM & 36.6 & 25.1 \\
         & \small MCTformer \cite{Xu2022MulticlassTT} \tiny 2022 & 58.8 & 50.5\\
          & \small Sparse ViT \cite{Hanna2023SparseMV} \tiny 2023 & 57.3 & 52.2 \\
          \rowcolor{gainsboro}
         & \small Ours & \underline{\textbf{74.1}} & \underline{\textbf{67.2}} \\
         \bottomrule
         & \multicolumn{3}{l}{\textbf{Supervised Methods}} \\
         \multirow{7}{*}{\rotatebox[origin=c]{90}{\small Endotect}} & \small UNet  & \underline{80.3} & 73.0 \\
         & \small ViT + seg. & 67.8 & 60.0 \\
         & \small Tomar \etal \cite{Tomar2020DDANetDD} \tiny 2020 & - & \underline{78} \\
         \cmidrule(lr){2-4}
         & \multicolumn{3}{l}{\textbf{Weakly Supervised Methods}} \\
         & \small ViT + GradCAM & 63.2 & 55.9\\
         & \small MCTformer \cite{Xu2022MulticlassTT} \tiny 2022 & 71.7& 63.3\\
         & \small Sparse ViT \cite{Hanna2023SparseMV} \tiny 2023 & 76.0& 68.0\\
         \rowcolor{gainsboro}
         & \small Ours  & \textbf{78.4} & \textbf{69.8} \\
         \bottomrule
         & \multicolumn{3}{l}{\textbf{Supervised Methods}} \\
         \multirow{7}{*}{\rotatebox[origin=c]{90}{\small ADE20K}}  & \small UNet  & 64.3 & 55.0 \\
         & \small ViT + seg.  & 43.5 & 32.2 \\
         & \small Zhou \etal \cite{Zhou2017ScenePT} \tiny 2017& \underline{74} & \underline{61} \\
         \cmidrule(lr){2-4}
         & \multicolumn{3}{l}{\textbf{Weakly Supervised Methods}} \\
         & \small ViT + GradCAM & 32.7 & 21.3\\
         & \small MCTformer \cite{Xu2022MulticlassTT} \tiny 2022 & 45.8 & 33.3\\
         & \small Sparse ViT \cite{Hanna2023SparseMV} \tiny 2023 & 44.2 & 34.5\\

         \rowcolor{gainsboro}
         & \small Ours & \textbf{51.8} & \textbf{38.2} \\
         \bottomrule
    \end{tabular}}
    \label{tab:baselines_specialized}
\end{table}

\begin{figure}
    \centering
    \includegraphics[width=\linewidth]{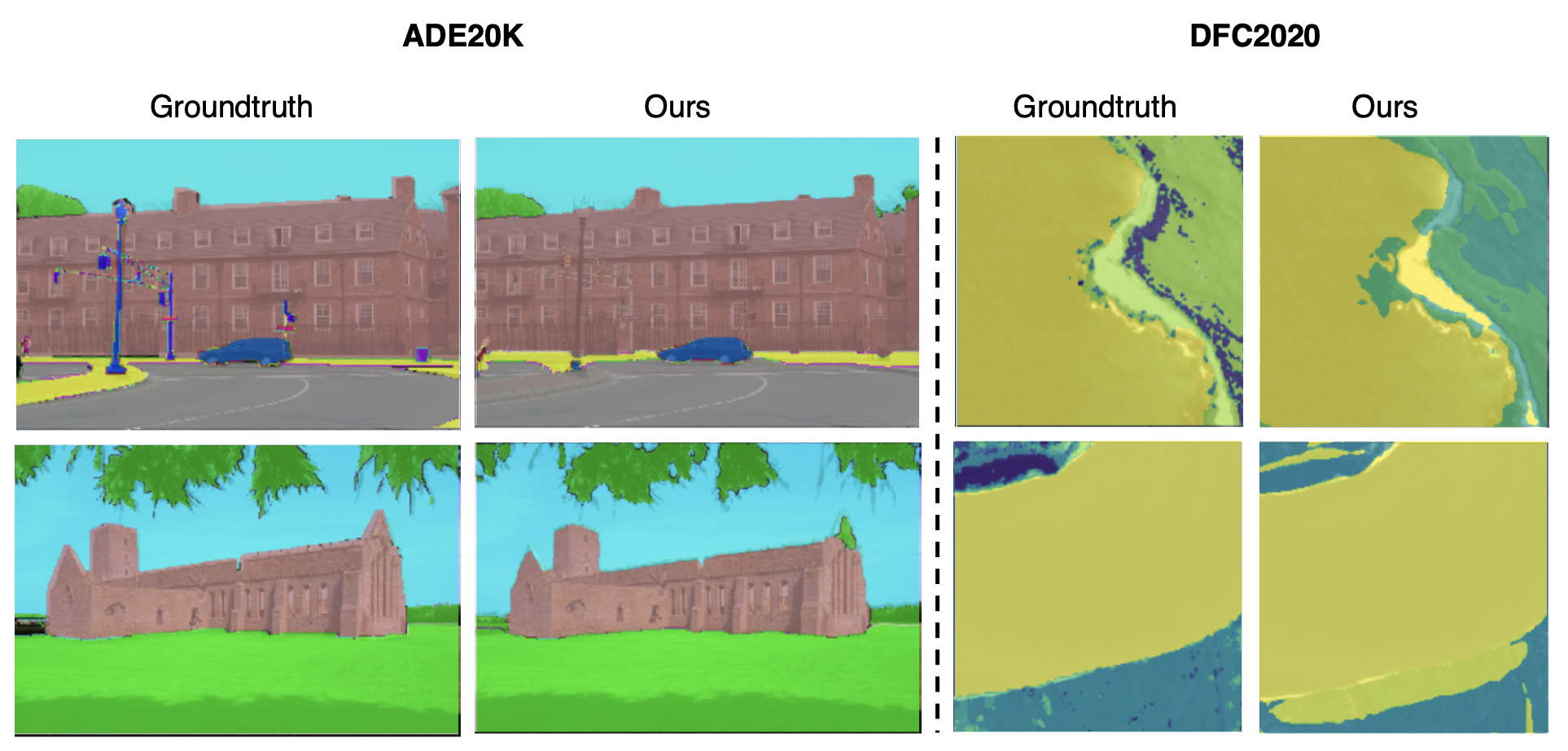}
    \caption{Segmentation results on the ADE20K and the DFC2020 datasets.}
    \label{fig:ade_dfc_comp}
\end{figure}

\subsection{Ablation Studies}
\begin{figure*}
    \centering
 \includegraphics[width=\textwidth]{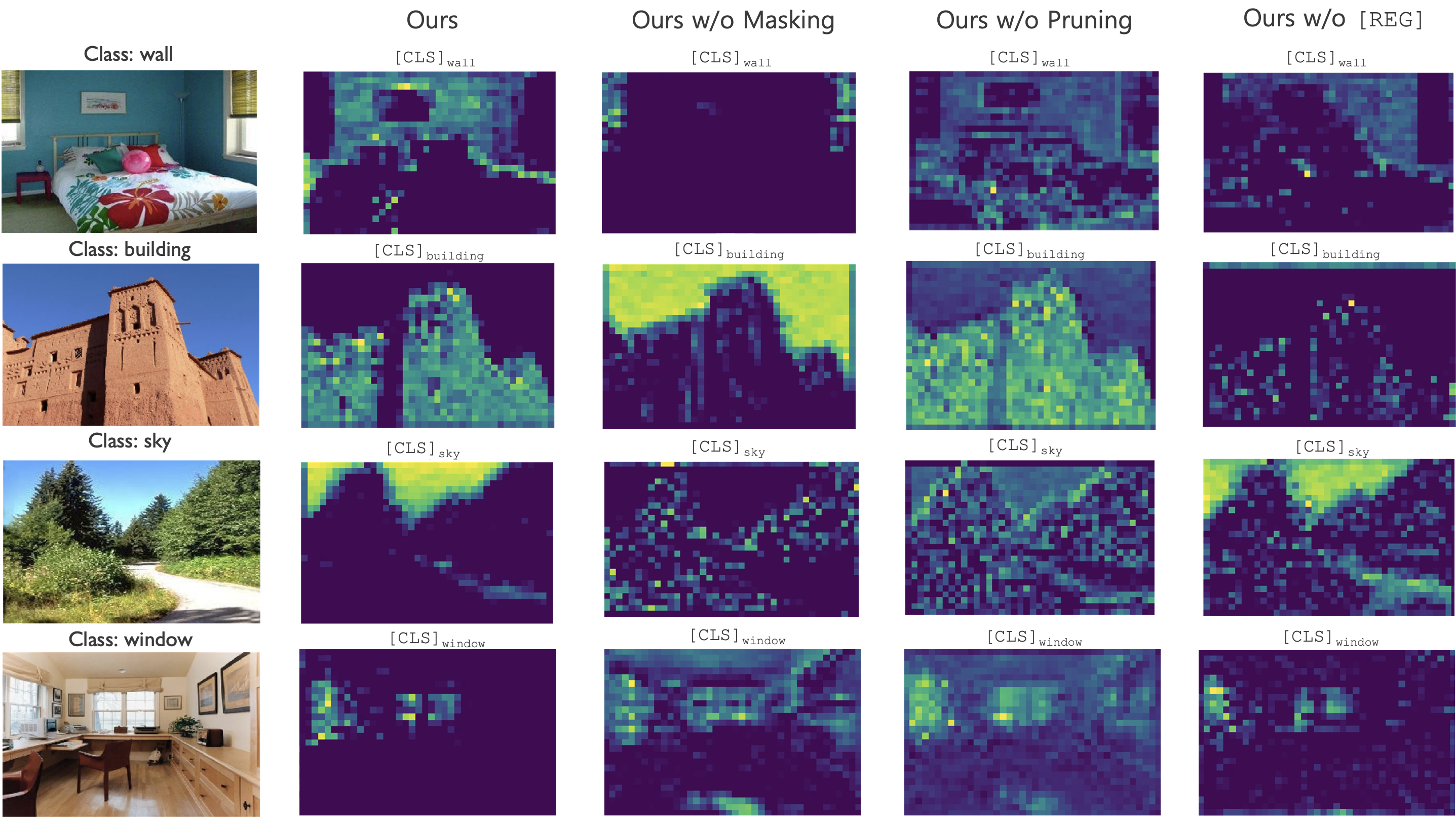}
    \caption{Visual comparison of the impact of different components on the self-attention of \texttt{[CLS]} tokens for the ADE20K dataset. For each image, one class is selected and the reshaped self-attention of the corresponding \texttt{[CLS]} token is displayed. Variations include (a) without random masking, (b) without head pruning, and (c) without appending the additional \texttt{[REG]} token.}
    \label{fig:cls_tokens}
    \vspace{-1em}
\end{figure*}

\begin{figure}[t]
    \centering
    \includegraphics[scale=0.24]{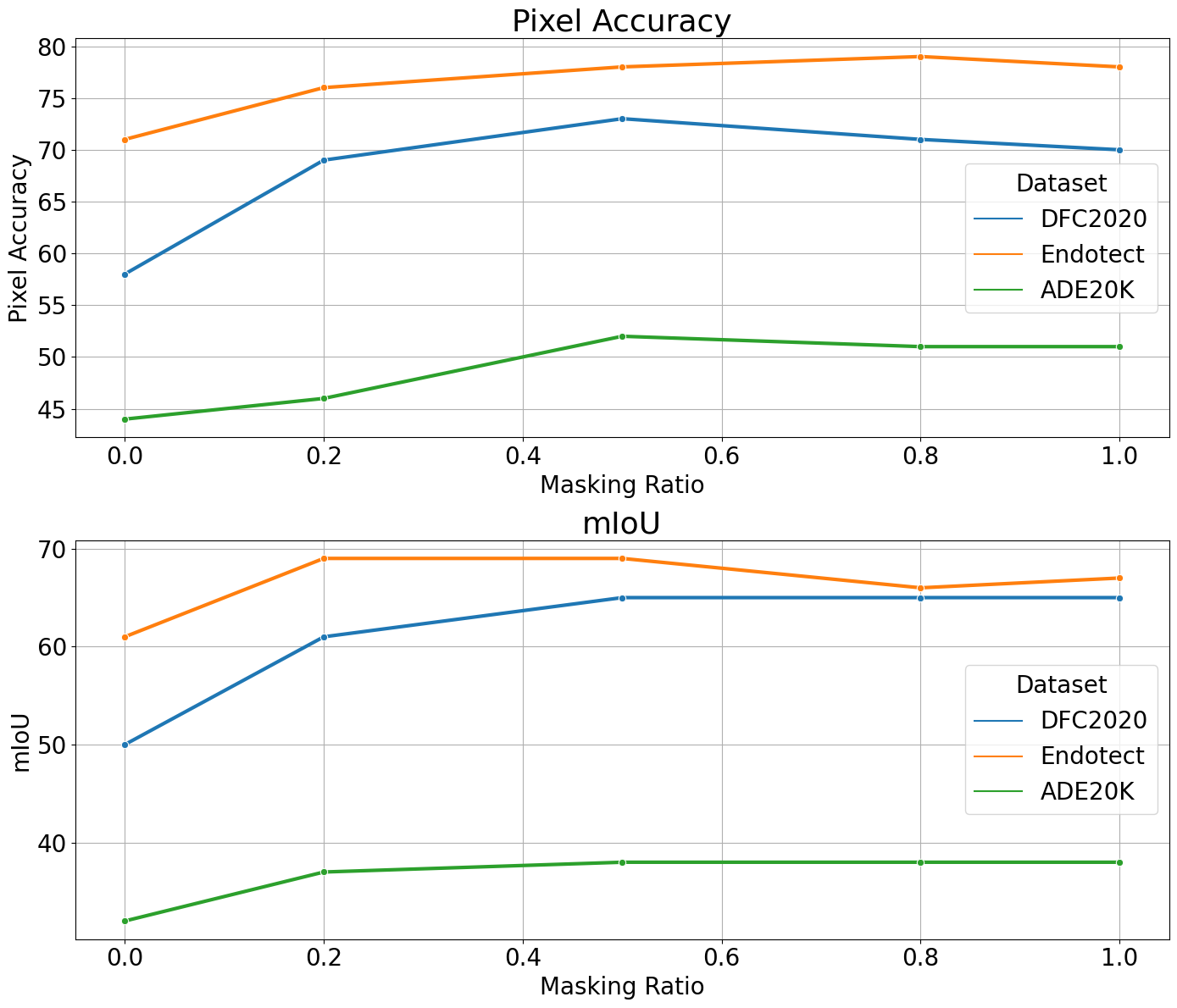}
    \caption{Masking ratios sensitivity analysis for the three domain-specific datasets. Results are reported for the generated pseudo-masks.}
    \label{fig:masking_ratio}
    % \vspace{-1.2em}
\end{figure}

To better understand the contribution of the different components in our method, we perform a series of ablation studies evaluating the effects of the random masking strategy, the additional \texttt{[REG]} token, and attention head pruning. The ablation studies are conducted across multiple datasets, with results summarized in both visualizations (Figure~\ref{fig:cls_tokens}) and quantitative tables (Tables~\ref{tab:random_masking}, \ref{tab:reg_token}, and \ref{tab:ah_pruning}).
\vspace{-1em}
\paragraph{Effect of the Random Masking Strategy}
We investigate the influence of the random masking strategy on the performance of our method (Section \ref{sec:method:random_masking}). The results are reported in Table \ref{tab:random_masking}. Randomly masking a subset of the \texttt{[CLS]} tokens is designed to ensure that each token is assigned one class to focus on. This strategy appears to be effective as indicated by the significant performance gap between the two setups, across both datasets. Without random masking, class assignments tend to be incorrect, leading to situations where the shapes are accurate, but the assigned classes are wrong, as observed in Figure~\ref{fig:cls_tokens}. In the column "Ours w/o Masking" the self-attention matrices fail to capture correct class-specific features, especially for "building", "sky", or "wall". A sensitivity analysis of the masking ratio is performed in Section \ref{sec:exp:sensitivity_analysis} and Figure \ref{fig:masking_ratio}.

\begin{table}
    \centering
    \caption{Effect of random (rnd) masking of a subset of the \texttt{[CLS]} tokens (50\%) on the pseudo-masks, with considerable improvement on all datasets.}
    \scalebox{1.0}{
    \begin{tabular}{ccc}
       & \textbf{w/o Rnd Masking} & \textbf{w/ Rnd Masking} \\
        \toprule
         MS COCO & 41.9 & \textbf{43.2} \\
         VOC & 71.6 & \textbf{72.7} \\
         \bottomrule
    \end{tabular}}
    \label{tab:random_masking}
    \vspace{-1em}
\end{table}

\paragraph{Effect of the additional register \textbf{\texttt{[REG]}} token}
We investigate the effect of incorporating an additional learnable token on our model's performance (Section \ref{sec:method:multiple_tokens}). Specifically, we compare two configurations: using C \texttt{[CLS]} tokens (one for each class) versus using the same C \texttt{[CLS]} tokens along with an additional \texttt{[REG]} token intended to capture general context. Our results (Table \ref{tab:reg_token}) show that incorporating an additional \texttt{[REG]} consistently improves results across datasets. This indicates that the \texttt{[REG]} token provides complementary information that aids in refining class-specific assignments. In Figure~\ref{fig:cls_tokens}, the "Ours w/o [REG]" column shows that the absence of the \texttt{[REG]} token results in less precise delineation of features in certain areas. For instance, the boundaries between "sky" and "trees" or "window" and "wall" appear fuzzier and less distinct.

\begin{table}[b]
    \centering
    \caption{Effect of the additional register token (\texttt{[REG]}), with visible improvement on both datasets}
    \scalebox{1.0}{
     \begin{tabular}{ccc}
       & \textbf{w/o \texttt{[REG]} token} & \textbf{w/ \texttt{[REG]} token} \\
        \toprule
         MS COCO & 42.8 & \textbf{43.2} \\
         VOC & 72.3  & \textbf{72.7} \\
         \bottomrule
    \end{tabular}}
    \label{tab:reg_token}
\end{table}

\paragraph{Effect of Attention Head Pruning}
We explore the impact of pruning attention heads on the performance and interpretability of our Vision Transformer model (Section \ref{sec:method:sparse_training}). Attention head pruning aims to discard redundant and noisy heads, potentially enhancing the focus of the attention mechanisms. Results are reported in Table \ref{tab:ah_pruning} and prove that pruning is indeed beneficial. Beyond accuracy, attention head pruning contributes to smoother pseudo-masks, as shown  in Figure~\ref{fig:cls_tokens}. The "Ours w/o Pruning" column reveals the consequences of retaining redundant attention heads. For example, in the "building" and "sky" classes, the masks appear noisy and fragmented, reflecting the influence of irrelevant attention patterns.

\begin{table}
    \centering
    \caption{Effect of attention head (AH) pruning on the generated pseudo-masks}
    \scalebox{1.0}{
     \begin{tabular}{ccc}
       & \textbf{w/o AH Pruning} & \textbf{w/ AH Pruning} \\
        \toprule
         MS COCO & 41.7 & \textbf{43.2} \\
         VOC & 72.0 & \textbf{72.7} \\
         \bottomrule
    \end{tabular}}
    \label{tab:ah_pruning}
\end{table}

\subsection{Sensitivity Analysis}
\label{sec:exp:sensitivity_analysis}
In this section, we conduct a sensitivity analysis to understand the impact of different masking ratios on the performance of our Vision Transformer model as a classifier and on the accuracy of the generated pseudo-masks. The masking ratio determines the proportion of \texttt{[CLS]} tokens (corresponding to classes that are not in the label)  that are randomly masked during each training step. We experiment with the following masking ratios: 0\% (no masking), 20\%, 50\%, 80\% and 100\%. Results are reported in Figure \ref{fig:masking_ratio}. As shown in the figure, increasing the masking ratio generally improves both pixel accuracy and mIoU up to around 50\%, after which performance plateaus or slightly declines, indicating an optimal balance between masking and preserving useful class information.

%% file: 6_discussion.tex
\section{General Discussion}
\label{sec:discussion}
The main challenges of using attention maps directly for weakly supervised semantic segmentation without any additional tools are twofold: (i) ensuring that each self-attention map captures distinct object classes in an image and (ii) correctly assigning each of these maps to its corresponding class for pseudo-mask generation. Traditional WSSS methods often struggle with these aspects, leading to coarse or overlapping segmentation masks. Our approach tackles these challenges by leveraging a Vision Transformer architecture with multiple \texttt{[CLS]} tokens, where each token corresponds to a specific class. This structured approach encourages the model to develop class-specific representations and enhances interpretability.

A common observation across diverse datasets is that our proposed method consistently produces accurate and well-defined segmentation shapes. This improvement is largely attributable to three key architectural choices: the introduction of an additional \texttt{[REG]} token, which captures background and global context; the pruning of redundant attention heads, which sharpens class-specific feature extraction; and the masking strategy, which ensures better class-token assignments by reducing interference between different class representations. These refinements contribute to the generation of high-quality pseudo-masks, which serve as strong initial segmentation candidates.

Another key strength of our approach is its effectiveness in multilabel segmentation tasks, such as ADE20K, where multiple objects coexist within a single image. Unlike single-label segmentation, which relies on coarse feature localization, multilabel segmentation demands precise class-token assignments. Our approach performs very well in these complex scenarios, demonstrating that the combination of multiple \texttt{[CLS]} tokens, class-aware masking, and structured attention regularization can lead to robust and interpretable segmentation results. 

However, an inherent trade-off emerges: as the number of classes (and consequently, the number of \texttt{[CLS]} tokens) grows, so does the model's parameter count. This results in increased computational complexity and memory requirements, posing scalability concerns. While our attention head pruning strategy mitigates some of these burdens by reducing redundancy, optimizing parameter efficiency remains an open challenge. Future research could explore dynamic token assignment strategies, reducing unnecessary computations while maintaining segmentation accuracy. Despite these computational challenges, our results demonstrate that this approach is highly effective when adequate resources are available.

%% file: 7_conclusion.tex
\section{Conclusion}
\label{sec:conclusion}
To conclude, this works introduces a novel single-stage weakly supervised segmentation method that effectively leverages the self-attention mechanisms of Vision Transformers. By incorporating multiple \texttt{[CLS]} tokens, applying random masking, and pruning redundant attention heads, we generate class-specific attention maps that translate into highly accurate pseudo-masks. Our extensive experiments across multiple datasets - including remote sensing, medical imaging, and general scene understanding - demonstrate the robustness and generalization of our approach. Our results show that our method significantly reduces the dependency on fine-grained labels while achieving performance comparable to fully supervised models.
As the demand for efficient and scalable segmentation solutions grows, our approach offers a step toward reducing annotation overhead while maintaining high segmentation quality. We believe that further advancements in weak supervision and attention-based learning will continue to push the boundaries of semantic segmentation research.